\documentclass[11pt]{article}
\usepackage{acl2002}
\usepackage{times}
\usepackage{latexsym}
\title{Assessing System Agreement and Instance Difficulty\\
in the Lexical Sample Tasks of \textsc{Senseval-2}}
\author{
Ted Pedersen\\
Department of Computer Science\\
University of Minnesota\\
Duluth, MN, 55812 USA\\
{\tt tpederse@d.umn.edu}
}

\date{}

\begin{document}
\maketitle
\begin{abstract}
This paper presents a comparative evaluation among the
systems that participated in the Spanish and English lexical sample
tasks of \textsc{Senseval-2}. The focus is on pairwise comparisons
among systems to assess the degree to which they agree, and on 
measuring the difficulty of the test instances included in 
these tasks.
\end{abstract}   

\section{Introduction}

 
This paper presents a post-mortem analysis of the English and Spanish  
lexical sample tasks of \textsc{Senseval-2}. Two closely related questions
are considered. First, to what extent did the competing systems agree?
Did systems tend to be redundant and have success with many of the same
test instances, or were they complementary and able to disambiguate   
different portions of the instance space? Second, how much did the 
difficulty of the test instances vary? Are there test instances that 
proved unusually difficult to disambiguate relative to other instances?

We address the first question via a series of pairwise comparisons among  
the participating systems that measures their agreement via the kappa  
statistic. We also introduce a simple measure of the degree to which 
systems are complementary called {\it optimal combination}. We analyze  
the second question by rating the difficulty of test instances  
relative to the number of systems that were able to disambiguate them  
correctly.  

Nearly all systems that received official scores in the Spanish and  
English lexical sample tasks of \textsc{Senseval-2} are included in this  
study. There are 23 systems included from the English lexical sample 
task and eight from the Spanish. Table \ref{tab:sum} lists the systems 
and shows the number of test instances that each disambiguated correctly, 
both by part of speech and in total.

\begin{table}
\caption{Lexical Sample Systems}
\begin{tabular}{lrrrr}
\hline
\hline
\multicolumn{1} {c} {system} &
\multicolumn{4} {c} {correct instances} \\
\multicolumn{1} {c} {name} &
\multicolumn{1} {c} {noun} &
\multicolumn{1} {c} {verb} &
\multicolumn{1} {c} {adj} &
\multicolumn{1} {c} {total (\%)} \\
\hline
\multicolumn{5} {c} {English} \\
\hline
      & 1754 & 1806 & 768 & 4328 (1.00) \\ \\
jhu\_final & 1196 & 1022 & 562 & 2780 (0.64) \\ 
smuls & 1219 & 1016 & 528 & 2763 (0.64) \\ 
kunlp & 1171 & 1040 & 513 & 2724 (0.63) \\ 
cs224n & 1198 & 945 & 527 & 2670 (0.62) \\ 
lia & 1177 & 966 & 510 & 2653 (0.61) \\ 
talp & 1149 & 927 & 495 & 2571 (0.59) \\ 
duluth3 & 1137 & 840 & 497 & 2473 (0.57) \\ 
umcp & 1081 & 891 & 487 & 2459 (0.57) \\ 
ehu\_all & 1069 & 891 & 480 & 2440 (0.56) \\ 
duluth4 & 1065 & 806 & 476 & 2346 (0.54) \\ 
duluth2 & 1056 & 795 & 483 & 2334 (0.54) \\ 
lesk\_corp & 960 & 804 & 454 & 2218 (0.51) \\ 
duluthB & 1004 & 729 & 467 & 2200 (0.51) \\ 
uned\_ls\_t & 987 & 699 & 469 & 2155 (0.50) \\ 
common & 880 & 728 & 453 & 2061 (0.48) \\ 
alicante & 427 & 866 & 486 & 1779 (0.41) \\ 
uned\_ls\_u & 781 & 519 & 437 & 1736 (0.40) \\ 
clr\_ls & 602 & 393 & 272 & 1267 (0.29) \\ 
iit2 & 541 & 348 & 166 & 1054 (0.24) \\ 
iit1 & 516 & 337 & 182 & 1034 (0.24) \\ 
lesk & 467 & 328 & 182 & 977 (0.23) \\ 
lesk\_def & 438 & 159 & 108 & 704 (0.16) \\ 
random & 303 & 153 & 155 & 611 (0.14) \\ 
\hline
\multicolumn{5} {c} {Spanish} \\ 
\hline 
      & 799 & 745 & 681 & 2225 (1.00)\\ \\
jhu & 560 & 478 & 546 & 1584 (0.71) \\ 
cs224 & 520 & 443 & 526 & 1489 (0.67) \\ 
umcp & 482 & 435 & 479 & 1396 (0.63) \\ 
duluth8 & 494 & 382 & 494 & 1369 (0.62) \\ 
duluth7 & 470 & 374 & 480 & 1324 (0.60) \\ 
duluth9 & 445 & 359 & 446 & 1250 (0.56) \\ 
duluthY & 411 & 325 & 434 & 1170 (0.53) \\ 
alicante & 269 & 381 & 468 & 1118 (0.50) \\ 
\end{tabular}
\label{tab:sum}
\end{table}

\section{Pairwise System Agreement}

Assessing agreement among systems sheds light on whether their combined  
performance is potentially more accurate than that of any of the 
individual systems. If several systems are largely in agreement, 
then there is little benefit in combining them since they are redundant  
and will simply reinforce one another. However, if some systems 
disambiguate instances that others do not, then they are complementary  
and it may be possible to combine them to take advantage of the different  
strengths of each system to improve overall accuracy. 

The kappa statistic \cite{Cohen60} is a measure of agreement between  
multiple systems (or judges) that is scaled by the agreement that would
be expected just by chance. A value of 1.00 suggests complete agreement,  
while 0.00 indicates pure chance agreement. Negative values indicate  
agreement less than what would be expected by chance.   
\cite{Krippendorf80} points out that it is   
difficult to specify a particular value of kappa as being generally  
indicative of agreement. As such we simply use kappa as a tool for  
comparison and relative ranking.  A detailed discussion on the use of  
kappa in natural language processing is presented in \cite{Carletta96}. 

To study agreement we have made a series of pairwise comparisons among  
the systems included in the English and Spanish lexical sample tasks. 
Each pairwise combination is represented in a 2 $\times$ 2    
contingency table, where one cell represents the number of test instances  
that both systems disambiguate correctly, one cell represents the number  
of instances where both systems are incorrect, and there are two  
cells to represent the counts when only one system is correct. 
Agreement does not imply accuracy, since two systems may get a large
number of the same instances incorrect and have a high rate of agreement.

Tables \ref{tab:eng-kappa} and \ref{tab:spa-kappa} show the system  
pairs in the English and Spanish lexical sample tasks that exhibit the  
highest level of agreement according to the kappa statistic. 
The values in the both--one--zero column indicate the percentage of  
instances where both systems are correct, where only one is correct, and  
where neither is correct. The top 15 pairs are shown for nouns and  verbs, 
and the top 10 for adjectives. 
A complete list would include about 250 pairs for each part of  
speech for English and 24 such pairs for Spanish. 


\begin{table} 
\caption{Pairwise Agreement English} 
\begin{tabular}{lcc}\hline \hline 
\multicolumn{1} {l} {system pair} & 
\multicolumn{1} {c} {both-one-zero} & 
\multicolumn{1} {c} {kappa} \\ 
\hline 
\multicolumn{3}{c} {Nouns} \\ 
common lesk\_corp & 0.49 0.06 0.44 & 0.87 \\ 
duluth2 duluth3 & 0.60 0.08 0.32 & 0.82 \\ 
lesk\_corp umcp & 0.53 0.11 0.36 & 0.78 \\ 
duluth2 duluthB & 0.54 0.14 0.32 & 0.70 \\ 
iit1 iit2 & 0.24 0.13 0.63 & 0.69 \\ 
duluth3 duluthB & 0.56 0.15 0.29 & 0.68 \\ 
common umcp & 0.48 0.16 0.36 & 0.68 \\ 
ehu\_all umcp & 0.55 0.17 0.29 & 0.64 \\ 
uned\_ls\_t uned\_ls\_u & 0.41 0.19 0.40 & 0.63 \\ 
duluth3 duluth4 & 0.55 0.18 0.27 & 0.61 \\ 
duluth2 duluth4 & 0.52 0.19 0.29 & 0.60 \\ 
duluth4 duluthB & 0.51 0.19 0.29 & 0.59 \\ 
ehu\_all lesk\_corp & 0.50 0.20 0.30 & 0.59 \\ 
cs224n duluth3 & 0.58 0.19 0.24 & 0.58 \\ 
cs224n duluth4 & 0.55 0.19 0.25 & 0.58\\ 
\hline 
\multicolumn{3} {c} {Verbs} \\ 
common lesk\_corp & 0.39 0.06 0.55 & 0.88 \\ 
duluth2 duluth3 & 0.43 0.07 0.50 & 0.85 \\ 
duluth3 duluth4 & 0.39 0.14 0.47 & 0.72 \\ 
duluth2 duluth4 & 0.38 0.15 0.47 & 0.69 \\
lesk\_corp umcp & 0.38 0.17 0.44 & 0.65 \\ 
common umcp & 0.36 0.18 0.46 & 0.65 \\ 
cs224n duluth3 & 0.40 0.20 0.40 & 0.60 \\ 
cs224n duluth4 & 0.38 0.21 0.41 & 0.59 \\
cs224n duluth2 & 0.38 0.21 0.40 & 0.57 \\ 
uned\_ls\_t uned\_ls\_u & 0.24 0.20 0.55 & 0.56 \\ 
duluth3 lia & 0.39 0.23 0.38 & 0.54 \\ 
lesk\_corp talp & 0.36 0.23 0.40 & 0.54 \\ 
cs224n lia & 0.41 0.24 0.35 & 0.53 \\ 
common talp & 0.34 0.24 0.42 & 0.52 \\ 
kunlp talp & 0.42 0.24 0.33 & 0.51 \\ 
\hline
\multicolumn{3} {c} {Adjectives} \\
common lesk\_corp & 0.59 0.00 0.41 & 0.99 \\ 
duluth2 duluth3 & 0.63 0.03 0.34 & 0.93 \\ 
lesk\_corp umcp & 0.58 0.07 0.35 & 0.86 \\ 
duluth2 duluthB & 0.59 0.07 0.35 & 0.86 \\ 
duluth3 duluthB & 0.60 0.07 0.34 & 0.86 \\ 
common umcp & 0.58 0.07 0.35 & 0.86 \\
duluth4 duluthB & 0.55 0.14 0.32 & 0.71 \\ 
duluth3 duluth4 & 0.57 0.14 0.30 & 0.71 \\ 
cs224n duluth3 & 0.60 0.13 0.27 & 0.70 \\ 
cs224n duluth2 & 0.59 0.14 0.27 & 0.70 \\ 
\end{tabular} 
\label{tab:eng-kappa}
\end{table}

\begin{table}
\caption{Pairwise Agreement Spanish}
\begin{tabular}{lcc}
\hline
\hline
\multicolumn{1} {l} {system pair} &
\multicolumn{1} {c} {both-one-zero} &
\multicolumn{1} {c} {kappa} \\
\hline
\multicolumn{3} {c} {Nouns} \\ 
duluth7 duluth8 & 0.57  0.11  0.32 & 0.76 \\
umcp duluth9 & 0.50  0.17  0.33 & 0.65 \\
duluth7 duluthY & 0.49  0.21  0.30 & 0.57 \\
umcp duluthY & 0.48  0.21  0.31 & 0.56 \\
duluth8 duluthY & 0.50  0.21  0.29 & 0.56 \\
umcp duluth8 & 0.50  0.23  0.27 & 0.51 \\
umcp duluth7 & 0.50  0.23  0.27 & 0.51 \\
cs224 umcp & 0.51  0.24  0.25 & 0.49 \\
duluth9 duluthY & 0.44  0.26  0.30 & 0.47 \\
duluth8 duluth9 & 0.47  0.27  0.27 & 0.45 \\
cs224 duluth9 & 0.47  0.27  0.26 & 0.44 \\
cs224 jhu & 0.55  0.25  0.20 & 0.43 \\
cs224 duluth8 & 0.51  0.27  0.22 & 0.42 \\
jhu umcp & 0.51  0.29  0.21 & 0.38 \\
jhu duluth8 & 0.53  0.28  0.19 & 0.37 \\
\hline
\multicolumn{3} {c} {Verbs} \\ 
duluth7 duluth8 & 0.48  0.08  0.44 & 0.84 \\
duluth8 duluth9 & 0.44  0.14  0.42 & 0.72 \\
umcp duluth8 & 0.48  0.14  0.37 & 0.71 \\
umcp duluth9 & 0.46  0.16  0.38 & 0.69 \\
duluth7 duluth9 & 0.42  0.16  0.42 & 0.68 \\
umcp duluth7 & 0.47  0.16  0.37 & 0.68 \\
duluth8 duluthY & 0.44  0.16  0.39 & 0.67 \\
duluth9 duluthY & 0.42  0.18  0.40 & 0.65 \\
duluth7 duluthY & 0.43  0.18  0.39 & 0.64 \\
umcp duluthY & 0.46  0.19  0.35 & 0.61 \\
cs224 umcp & 0.49  0.19  0.32 & 0.61 \\
cs224 duluth8 & 0.44  0.25  0.32 & 0.50 \\
alicante umcp & 0.42  0.25  0.33 & 0.50 \\
cs224 duluth7 & 0.43  0.26  0.32 & 0.48 \\
cs224 jhu & 0.49  0.25  0.26 & 0.48 \\
\hline
\multicolumn{3} {c} {Adjectives} \\ 
duluth7 duluth8 & 0.69  0.06  0.25 & 0.85 \\
duluth7 duluthY & 0.60  0.14  0.26 & 0.68 \\
umcp duluthY & 0.60  0.15  0.26 & 0.67 \\
umcp duluth9 & 0.61  0.14  0.25 & 0.67 \\
duluth8 duluthY & 0.61  0.15  0.24 & 0.67 \\
umcp duluth8 & 0.64  0.15  0.21 & 0.64 \\
duluth9 duluthY & 0.56  0.18  0.26 & 0.60 \\
duluth8 duluth9 & 0.61  0.17  0.22 & 0.59 \\
umcp duluth7 & 0.62  0.17  0.21 & 0.59 \\
duluth7 duluth9 & 0.58  0.20  0.22 & 0.54 \\
\end{tabular}
\label{tab:spa-kappa}
\end{table}

The utility of kappa agreement is confirmed in that system pairs known to  
be very similar have correspondingly high measures. In Table  
\ref{tab:eng-kappa},  duluth2 and duluth3 exhibit a high kappa value for   
all parts of speech. This is expected since duluth3 is an ensemble  
approach that includes duluth2 as one of its members. The same 
relationship exists between duluth7 and duluth8 in the Spanish lexical 
sample, and comparable behavior is seen in Table \ref{tab:spa-kappa}. 


A more surprising case is the even higher level of agreement  
between the most common sense baseline and the lesk 
corpus baseline shown in Table 2. This is  not necessarily  expected, and  
suggests that lesk corpus may not be finding a significant number of 
matches between the Senseval contexts and the WordNet glosses (as the  
lesk algorithm would hope to do) but instead may be relying on a 
simple default in many cases.

In previous work \cite{Pedersen01c} we propose a 50-25-25 rule that   
suggests that about half of the instances in a supervised word sense  
disambiguation evaluation will be fairly easy for most systems to  
resolve, another quarter will be harder but possible for at least some  
systems, and that the final quarter will be very difficult for any 
system to resolve. This same idea could also be expressed by stating 
that the kappa agreement between two word sense disambiguation  
systems will likely be around 0.50. In fact this is a common result
in the full set of pairwise comparisons, particularly for overall results
not broken down by part of speech. Tables 2 and 3 only list the largest  
kappa values,  but even there kappa quickly reduces towards 0.50. These 
same tables show that it is rare for two systems to agree on more than  
60\% of the correctly disambiguated instances. 

\begin{table}
\caption{Optimal Combination English}
\begin{tabular}{lcc}
\hline
\hline
\multicolumn{1} {l} {system pair} &
\multicolumn{1} {c} {both-one-zero} &
\multicolumn{1} {c} {kappa} \\
\hline
\multicolumn{3} {c} {Nouns} \\ 
kunlp smuls & 0.49  0.39  0.12 & 0.11 \\
smuls talp & 0.48  0.39  0.13 & 0.11 \\
cs224n kunlp & 0.48  0.39  0.13 & 0.11 \\
ehu\_all smuls & 0.47  0.40  0.13 & 0.10 \\
cs224n talp & 0.48  0.39  0.14 & 0.13 \\
jhu\_final kunlp & 0.49  0.37  0.14 & 0.16 \\
smuls umcp & 0.45  0.41  0.14 & 0.10 \\
kunlp lia & 0.48  0.38  0.14 & 0.14 \\
lia talp & 0.47  0.39  0.14 & 0.13 \\
jhu\_final talp & 0.48  0.38  0.14 & 0.15 \\
duluth3 kunlp & 0.47  0.38  0.15 & 0.15 \\
cs224n ehu\_all & 0.48  0.38  0.15 & 0.16 \\
ehu\_all lia & 0.47  0.38  0.15 & 0.15 \\
ehu\_all jhu\_final & 0.48  0.36  0.16 & 0.19 \\
duluth3 talp & 0.47  0.38  0.16 & 0.17 \\
\hline
\multicolumn{3} {c} {Verbs} \\ 
jhu\_final kunlp & 0.34  0.46  0.20 & 0.06 \\
ehu\_all jhu\_final & 0.31  0.44  0.21 & 0.07 \\
ehu\_all smuls & 0.31  0.44  0.21 & 0.07 \\
ehu\_all kunlp & 0.33  0.41  0.22 & 0.13 \\
kunlp smuls & 0.36  0.43  0.22 & 0.13 \\
cs224n ehu\_all & 0.29  0.45  0.22 & 0.05 \\
ehu\_all lia & 0.30  0.44  0.22 & 0.08 \\
cs224n kunlp & 0.33  0.44  0.23 & 0.11 \\
alicante ehu\_all & 0.26  0.47  0.23 & 0.03 \\
kunlp lia & 0.34  0.42  0.23 & 0.14 \\
jhu\_final talp & 0.32  0.44  0.24 & 0.12 \\
duluth3 ehu\_all & 0.26  0.46  0.24 & 0.05 \\
ehu\_all talp & 0.30  0.41  0.24 & 0.13 \\
alicante jhu\_final & 0.30  0.45  0.24 & 0.09 \\
jhu\_final umcp & 0.30  0.45  0.24 & 0.09 \\
\hline
\multicolumn{3} {c} {Adjectives} \\ 
alicante jhu\_final & 0.46  0.37  0.08 & 0.03 \\
alicante smuls & 0.41  0.41  0.09 & -0.04 \\
alicante cs224n & 0.42  0.40  0.09 & -0.01 \\
alicante kunlp & 0.41  0.39  0.11 & 0.03 \\
alicante lia & 0.41  0.39  0.11 & 0.03 \\
alicante duluth3 & 0.40  0.40  0.11 & 0.02 \\
alicante talp & 0.40  0.41  0.11 & 0.02 \\
alicante ehu\_all & 0.41  0.39  0.11 & 0.05 \\
alicante umcp & 0.39  0.40  0.12 & 0.04 \\
alicante duluth2 & 0.39  0.40  0.12 & 0.03 \\
\end{tabular}
\label{tab:eng-opt}
\end{table}

\begin{table}
\caption{Optimal Combination Spanish}
\begin{tabular}{lcc}
\hline
\hline
\multicolumn{1} {l} {system pair} &
\multicolumn{1} {c} {both-one-zero} &
\multicolumn{1} {c} {kappa} \\
\hline
\multicolumn{3} {c} {Nouns} \\ 
alicante jhu & 0.29  0.32  0.11 & 0.06 \\
alicante duluth7 & 0.27  0.34  0.12 & 0.03 \\
alicante duluthY & 0.25  0.35  0.12 & 0.01 \\
alicante duluth8 & 0.28  0.32  0.13 & 0.08 \\
alicante cs224 & 0.28  0.32  0.13 & 0.09 \\
alicante umcp & 0.26  0.33  0.14 & 0.06 \\
alicante duluth9 & 0.26  0.31  0.16 & 0.14 \\
jhu duluthY & 0.46  0.36  0.18 & 0.24 \\
jhu duluth7 & 0.51  0.29  0.19 & 0.35 \\
jhu duluth8 & 0.53  0.28  0.19 & 0.37 \\
cs224 jhu & 0.55  0.25  0.20 & 0.43 \\
jhu duluth9 & 0.46  0.34  0.20 & 0.29 \\
jhu umcp & 0.51  0.29  0.21 & 0.38 \\
cs224 duluth7 & 0.49  0.30  0.22 & 0.36 \\
cs224 duluth8 & 0.51  0.27  0.22 & 0.42 \\
\hline
\multicolumn{3} {c} {Verbs} \\ 
jhu duluthY & 0.39  0.38  0.23 & 0.23 \\
jhu umcp & 0.48  0.27  0.25 & 0.44 \\
jhu duluth9 & 0.39  0.36  0.26 & 0.29 \\
jhu duluth8 & 0.42  0.32  0.26 & 0.35 \\
cs224 jhu & 0.49  0.25  0.26 & 0.48 \\
jhu duluth7 & 0.42  0.32  0.26 & 0.36 \\
alicante jhu & 0.45  0.26  0.29 & 0.47 \\
cs224 duluthY & 0.43  0.27  0.30 & 0.46 \\
alicante cs224 & 0.41  0.28  0.31 & 0.44 \\
alicante duluthY & 0.35  0.34  0.31 & 0.32 \\
cs224 umcp & 0.49  0.19  0.32 & 0.61 \\
cs224 duluth7 & 0.43  0.26  0.32 & 0.48 \\
cs224 duluth8 & 0.44  0.25  0.32 & 0.50 \\
cs224 duluth9 & 0.41  0.27  0.32 & 0.47 \\
alicante umcp & 0.42  0.25  0.33 & 0.50 \\
\hline
\multicolumn{3} {c} {Adjectives} \\ 
jhu duluth8 & 0.66  0.22  0.12 & 0.39 \\
jhu duluth7 & 0.64  0.24  0.12 & 0.36 \\
jhu duluthY & 0.56  0.31  0.12 & 0.25 \\
alicante jhu & 0.62  0.26  0.13 & 0.33 \\
jhu duluth9 & 0.59  0.29  0.13 & 0.29 \\
cs224 jhu & 0.70  0.16  0.13 & 0.51 \\
jhu umcp & 0.64  0.23  0.13 & 0.38 \\
alicante cs224 & 0.61  0.24  0.15 & 0.39 \\
cs224 duluth8 & 0.66  0.19  0.16 & 0.50 \\
cs224 duluth7 & 0.64  0.20  0.16 & 0.49 \\
\end{tabular}
\label{tab:spa-opt}
\end{table}

\begin{table}
\caption{Difficulty of Instances}
\begin{tabular}{crrrr}
\hline
\hline
\multicolumn{1} {c} {\#} &
\multicolumn{1} {c} {noun} &
\multicolumn{1} {c} {verb} &
\multicolumn{1} {c} {adj} &
\multicolumn{1} {c} {total} \\
\hline
\multicolumn{5} {c} {English} \\ 
0  & 59 (16) & 174 (6) & 29 (8) & 262 (8) \\ 
1  & 51 (15) & 116 (10) & 26 (14) & 193 (12) \\ 
2  & 59 (18) & 122 (12) & 41 (21) & 222 (15) \\ 
3  & 64 (19) & 117 (16) & 29 (23) & 210 (18) \\ 
4  & 84 (17)  & 102 (16) & 28 (18) & 214 (17) \\ 
5  & 76 (23) & 76 (18) & 24 (20) & 176 (21) \\ 
6  & 53 (28) & 61 (30) & 23 (31) & 137 (29) \\ 
7  & 51 (29) & 65 (22) & 23 (34) & 139 (27) \\ 
8  & 62 (27) & 58 (34) & 18 (31) & 138 (30) \\ 
9  & 47 (32) & 69 (28) & 17 (26) & 133 (29) \\ 
10  & 62 (28) & 61 (32) & 18 (30) & 141 (30) \\ 
11  & 55 (39) & 56 (26) & 21 (38) & 132 (34) \\ 
12  & 80 (40) & 61 (41) & 22 (35) & 163 (40) \\ 
13  & 86 (58) & 56 (34) & 21 (45) & 163 (48) \\ 
14  & 125 (65) & 62 (49) & 33 (51) & 220 (59) \\ 
15  & 131 (77) & 125 (99) & 36 (60) & 292 (84) \\ 
16  & 141 (83) & 107 (117) & 61 (70) & 309 (92) \\ 
17  & 133 (75) & 100 (162) & 86 (74) & 319 (101) \\ 
18  & 92 (73) & 80 (203) & 102 (80) & 274 (113) \\ 
19  & 97 (68) & 59 (170) & 49 (77) & 205 (100) \\ 
20  & 65 (66) & 38 (192) & 30 (49) & 133 (96) \\ 
21  & 42 (68) & 15 (155) & 17 (47) & 74 (79) \\ 
22  & 29 (70) & 15 (73) & 7 (39) & 51 (67) \\ 
23  & 10 (49) & 11 (52) & 7 (38) & 28 (47) \\ 
\hline
\multicolumn{5} {c} {Spanish} \\ 
0  & 50 (16) & 126 (12) & 52 (24) & 228 (16) \\ 
1  & 81 (18) & 63 (17) & 32 (36) & 176 (21) \\ 
2  & 63 (24) & 69 (18) & 42 (50) & 174 (28) \\ 
3  & 63 (27) & 55 (23) & 39 (81) & 157 (39) \\ 
4  & 74 (32) & 47 (23) & 43 (101) & 164 (47) \\ 
5  & 94 (35) & 49 (28) & 35 (77) & 178 (42) \\ 
6  & 87 (40) & 61 (39) & 57 (90) & 205 (53) \\ 
7  & 182 (47) & 94 (46) & 88 (93) & 364 (58) \\ 
8  & 105 (44) & 181 (62) & 293 (166) & 579 (111) \\ 
\end{tabular}
\label{tab:instbytrain}
\end{table}

\begin{table}
\caption{Difficulty of English Word Types}
\begin{tabular}{lr|lr}
\hline
\hline
\multicolumn{1} {l} {word-pos (test)} &
\multicolumn{1} {c} {mean} &
\multicolumn{1} {l} {word-pos (test)} &
\multicolumn{1} {c} {mean} \\
\hline
collaborate-v  (30) & 20.2  & circuit-n  (85) & 10.7  \\ 
solemn-a  (25) & 18.3  & sense-n  (53) & 10.6  \\ 
holiday-n  (31) & 17.7  & authority-n  (92) & 10.5  \\ 
dyke-n  (28) & 17.5  & replace-v  (45) & 10.4  \\ 
graceful-a  (29) & 17.3 & restraint-n  (45) & 10.3  \\ 
vital-a  (38) & 16.7  & live-v  (67) & 10.2  \\ 
detention-n  (32) & 16.5  & treat-v  (44) & 10.1  \\ 
faithful-a  (23) & 16.5  & free-a  (82) & 10.0  \\ 
yew-n  (28) & 16.1  & nature-n  (46) & 10.0  \\ 
chair-n  (69) & 16.0  & simple-a  (66) &  9.8  \\ 
ferret-v  (1) & 16.0  & dress-v  (59) &  9.7  \\ 
blind-a  (55) & 15.7  & cool-a  (52) &  9.7  \\ 
lady-n  (53) & 15.5  & bar-n  (151) &  9.5  \\ 
spade-n  (33) & 15.3  & stress-n  (39) &  9.5  \\ 
hearth-n  (32) & 15.1  & channel-n  (73) &  9.2  \\ 
face-v  (93) & 15.1  & match-v  (42) &  9.0  \\ 
green-a  (94) & 14.9  & natural-a  (103) &  9.0  \\ 
fatigue-n  (43) & 14.9  & serve-v  (51) &  8.8  \\ 
oblique-a  (29) & 14.3  & train-v  (63) &  8.7  \\ 
nation-n  (37) & 14.0  & post-n  (79) &  8.7  \\ 
church-n  (64) & 13.8  & fine-a  (70) &  8.6  \\ 
local-a  (38) & 13.6  & drift-v  (32) &  7.7  \\ 
fit-a  (29) & 13.4  & leave-v  (66) &  7.7  \\ 
use-v  (76) & 13.4  & play-v  (66) &  7.5  \\ 
child-n  (64) & 13.0  & wash-v  (12) &  7.4  \\ 
wander-v  (50) & 12.9  & keep-v  (67) &  7.4  \\ 
begin-v  (280) & 12.6  & work-v  (60) &  7.0  \\ 
bum-n  (45) & 12.5  & drive-v  (42) &  6.8  \\ 
feeling-n  (51) & 11.4  & develop-v  (69) &  6.6  \\ 
facility-n  (58) & 11.1  & carry-v  (66) &  6.3  \\ 
colorless   (35) & 11.1  &  see-v  (69) &  6.3  \\ 
grip-n  (51) & 11.1  & strike-v  (54) &  5.9  \\ 
day-n  (145) & 11.0  & call-v  (66) &  5.8  \\ 
mouth-n  (60) & 11.0  & pull-v  (60) &  5.7  \\ 
material-n  (69) & 11.0  & turn-v  (67) &  5.0  \\ 
art-n  (98) & 10.7  & draw-v  (41) &  4.7  \\ 
          &       & find-v  (68) &  4.2  \\ 
\normalsize
\end{tabular}
\label{tab:eng-words}
\end{table}

\begin{table}
\caption{Difficulty of Spanish Word Types}
\begin{tabular}{lr|lr}
\hline
\hline
\multicolumn{1} {l} {word-pos (test)} &
\multicolumn{1} {c} {mean} &
\multicolumn{1} {l} {word-pos (test)} &
\multicolumn{1} {c} {mean} \\
\hline
claro-a  (66) &  7.6 & verde-a  (33) &  5.3 \\
local-a  (55) &  7.4 & canal-n  (41) &  5.3 \\
popular-a  (204) &  7.1 & clavar-v  (44) &  5.1 \\
partido-n  (57) &  7.0 &  masa-n  (41) &  5.1 \\
bomba-n  (37) &  6.8 & apuntar-v  (49) &  4.9 \\
brillante-a  (87) &  6.7 & autoridad-n  (34) &  4.9 \\
usar-v  (56) &  6.5 &  tocar-v  (74) &  4.8 \\
tabla-n  (41) &  6.3 &  explotar-v  (41) &  4.7 \\
vencer-v  (65) &  6.3 & programa-n  (47) &  4.7 \\
simple-a  (57) &  6.2 & circuito-n  (49) &  4.3 \\
hermano-n  (57) &  6.1 & copiar-v  (53) &  4.3 \\
apoyar-v  (73) &  6.0 & actuar-v  (55) &  4.2 \\
vital-a  (79) &  5.9 & operacion-n  (47) &  4.2 \\
gracia-n  (61) &  5.9 & pasaje-n  (41) &  4.1 \\
organo-n  (81) &  5.8 & saltar-v  (37) &  4.1 \\
corona-n  (40) &  5.5 & tratar-v  (70) &  3.9 \\
ciego-a  (42) &  5.5 &  natural-a  (58) &  3.9 \\
corazon-n  (47) &  5.5 & grano-n  (22) &  3.9 \\
coronar-v  (74) &  5.4 & conducir-v  (54) &  3.8 \\ 
naturaleza-n  (56) &  5.4 & & \\
\end{tabular}
\label{tab:spa-words}
\end{table}

\section {Optimal Combination}

An {\it optimal combination} is the accuracy that could be attained by a   
hypothetical tool called an {\it optimal combiner} that accepts as 
input the sense assignments for a test instance as generated by several  
different systems. It is able to select the correct sense from these  
inputs, and will only be wrong when none of the sense assignments is the 
correct one. 
Thus, the percentage accuracy of an optimal combiner is equal to one  
minus the percentage of instances that no system can resolve correctly.  

Of course this is only a tool for thought experiments and is not a  
practical algorithm. An optimal combiner can establish an upper bound 
on the accuracy that could reasonably be attained over a particular  
sample of test instances.   

Tables \ref{tab:eng-opt} and \ref{tab:spa-opt} list the top system pairs  
ranked by optimal combination (1.00 - value in zero column) for the  
English and Spanish lexical samples. Kappa scores are also shown to 
illustrate the interaction between agreement and optimal combination. 
Optimal  combination is  maximized when the percentage of instances  where  
both systems are wrong is minimized. Kappa  agreement is maximized by  
minimizing the percentage of instances  where  one or the other system  
(but not both) is correct. Thus, the only way a system pair could have a  
high measure of kappa and a high measure of optimal combination is if  
they were very accurate systems that disambiguated many of the same test  
instances correctly.

System pairs with low measures of agreement are potentially  
quite interesting because they are the most likely to make complementary
errors. For example, in Table \ref{tab:spa-opt} under nouns, the  
alicante system has a low level of agreement with all of the other  
systems. However, the measure of optimal combination is quite high,  
reaching 0.89 (1.00 - 0.11) for the pair of alicante and jhu. In
fact, all seven of the other systems achieve their highest optimal 
combination value when paired with alicante. 

This combination of circumstances suggests that the alicante system is  
fundamentally different than the other systems, and is able to  
disambiguate a certain set of instances where the other systems fail. In   
fact the alicante system is different in that it is the only Spanish  
lexical sample system that makes use of the structure of Euro-WordNet,  
the source of the sense inventory. 


\section{Instance Difficulty}

The difficulty of disambiguating word senses can vary considerably. A word  
with multiple closely related senses is likely to be more difficult than  
one with a few starkly drawn differences. In supervised learning, a   
particular sense of a word can be difficult to disambiguate if there are a  
small number of training examples available. 

Table \ref{tab:instbytrain} shows the distribution of the number of  
instances that are  successfully disambiguated by a particular number of  
systems in both the English and Spanish lexical samples. The value   
under the \# column shows the number of systems that are able to  
disambiguate the number of noun, verb, adjective and total instances  
shown in the row. The average number of training examples available for  
the correct answers associated with these instances is shown in   
parenthesis. For example, the first line shows that there were 59 noun  
instances that no system (of 23) could disambiguate, and that there were  
on average 16 training examples available for each of the correct senses  
for these 59 instances. 

Two very clear trends emerge. First, there are a substantial number  of  
instances that are not disambiguated correctly by any system (262 in   
English, 228 in Spanish) and there are a large number of instances that  
are disambiguated by just a handful of systems. In the English lexical  
sample, there are 1,277 test instances that are correctly disambiguated by  
five or fewer of the 23 systems. This is nearly 30\% of the test data,  
and confirms that this was a very challenging set of test instances.  

There is also a very clear correlation between the number of training  
examples available for a particular sense of a word and the number of  
systems that are able to disambiguate instances of that word correctly.  
For example, Table 6 shows that there were 174 English verb instances that  
no system disambiguated correctly. On average there were only 6 training  
examples for the  correct senses of these instances.  However, there were  
28 instances that all 23 English systems were able  to disambiguate. For  
these instances an average of 47 training examples were available for each 
correct sense.

This correlation between instance difficulty and number of training 
examples may suggest that future  \textsc{Senseval} exercises provide a  
minimum number of training  examples for each sense, or adjust the scoring  
to reflect the difficulty of  disambiguating a sense with very few  
training examples.  

Finally, we assess the difficulty associated with word types by  
calculating the average number of systems that were able to disambiguate  
the instances associated with that type. This information is provided for   
the English and Spanish lexical samples in Tables \ref{tab:eng-words} and  
\ref{tab:spa-words}. Each word is shown with its part of speech,  
the number of test instances, and the average number of systems that were 
able to  disambiguate each of the test instances.

The verb {\it collaborate} is the easiest according to this metric in the 
English lexical sample. It has 30 test instances that were disambiguated  
correctly by an average of 20.2 of the 23 systems. The verb {\it find}  
proves to be the most difficult, with 68 test instances disambiguated  
correctly by an average of 4.2 systems. A somewhat less extreme range of 
values is observed for the Spanish lexical sample in Table 
\ref{tab:spa-words}. The adjective  {\it claro} had 66 test instances that 
were disambiguated correctly by an  average of 7.6 of the 8 systems.  The  
most difficult word  was the verb {\it conducir}, which has 54 test 
instances   that were  disambiguated correctly by an average of 3.8 
systems. 

\section{Conclusion}

This paper presents an analysis of the results from  the English and   
Spanish lexical sample tasks of \textsc{Senseval-2}. The analysis is 
based on the kappa statistic and a measure known as optimal  
combination. It also assesses the difficulty of the test instances in 
these lexical samples. We find that there are a significant number
of test instances that were not disambiguated correctly by any system,
and that there is some correlation between instance difficulty and
the number of available training examples. 

\section{Acknowledgments}

This work has been partially supported by a National Science Foundation
Faculty Early CAREER Development award (\#0092784). 
                                   


\end{document}